\pdfoutput=1

\documentclass[11pt]{article}

\usepackage[preprint]{acl}

\usepackage{times}
\usepackage{latexsym}
\usepackage{tabularx}
\usepackage{enumitem}
\usepackage{makecell}
\usepackage[T1]{fontenc}

\usepackage[utf8]{inputenc}

\usepackage{microtype}
\usepackage{booktabs}
\usepackage{inconsolata}

\usepackage{graphicx}
\graphicspath{ {.assets/images/} }
\newcolumntype{b}{X}
\newcolumntype{s}{>{\hsize=.5\hsize}X}
\title{Exploring Robustness of LLMs to Paraphrasing Based on Sociodemographic Factors}

\author{Pulkit Arora\textsuperscript{$\clubsuit$} \and Akbar Karimi\textsuperscript{$\clubsuit\spadesuit$} \and Lucie Flek\textsuperscript{$\clubsuit\spadesuit$}\\
\vspace{-1mm}\\
\textsuperscript{$\clubsuit$}Conversational AI and Social Analytics (CAISA) Lab, University of Bonn\\
\textsuperscript{$\spadesuit$}Lamarr Institute for Machine Learning and Artificial Intelligence\\
\texttt{ak@bit.uni-bonn.de}}

\begin{document}

\maketitle

\begin{abstract}
Despite their linguistic prowess, LLMs have been shown to be vulnerable to small input perturbations. While robustness to local adversarial changes has been studied, robustness to global modifications such as different linguistic styles remains underexplored. Therefore, we take a broader approach to explore a wider range of variations across sociodemographic dimensions. We extend the SocialIQA dataset to create diverse paraphrased sets conditioned on sociodemographic factors (age and gender). The assessment aims to provide a deeper understanding of LLMs in (a) their capability of generating demographic paraphrases with engineered prompts and (b) their capabilities in interpreting real-world, complex language scenarios. We also perform a reliability analysis of the generated paraphrases looking into linguistic diversity and perplexity as well as manual evaluation.
We find that demographic-based paraphrasing significantly impacts the performance of language models, indicating that the subtleties of linguistic variation remain a significant challenge. We will make the code and dataset available for future research. 
\end{abstract} 

\section{Introduction}
In recent years, Large Language Models (LLMs) like ChatGPT have drastically improved in generating human-like text, especially in tasks like question answering \cite{tan2023evaluation}. However, concerns persist regarding their robustness to diverse linguistic variations inherently present in the language used by different demographic groups \cite{marx2023effects, wang2024large}.  
\begin{figure}[t]
\centering
\includegraphics[scale=0.39]{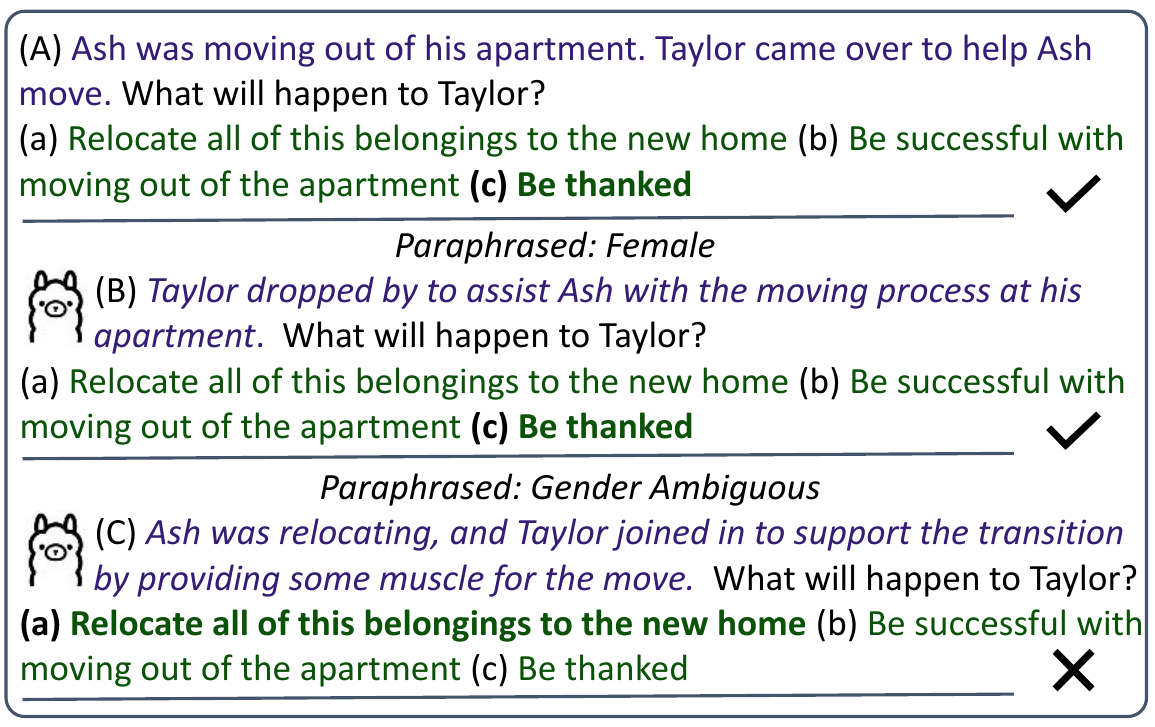}
\caption{Demographic paraphrasing by \texttt{LLAMA2-chat}. (A) A sample question from SocialIQA dataset (B) Female-style paraphrased context of the question (C) Gender-Ambiguous style paraphrased context of the question. LLM's predictions are \textbf{boldfaced}.}
\label{fig:paraphrase_example}
\end{figure}
Prior investigations, by frameworks like HELM \cite{bommasani2023holistic}, have assessed these models' capacity to manage specific rule-based text alterations and adversarial attacks. However, local adversarial attacks do not capture the linguistic diversity with which a text can be expressed. \citet{tan-etal-2021-reliability} defined a more structured approach to reliability testing through a DOCTOR framework, which proposes reliability tests to be dimension-specific worst-case test sets meant to mimic real-world applications. 
Previous robustness evaluations have not been focused on style variations across real-world language usage. This entails navigating through diverse linguistic variations, for instance, as shown in Figure \ref{fig:paraphrase_example}, a style change in paraphrasing of the original context produces an incorrect choice by the model.

We introduce linguistic diversity to the LLM's input context to address these issues and investigate their robustness to demographic paraphrases based on age and gender.  
Our results highlight that LLMs, despite being impressive generalizers of their training corpora, struggle to accurately interpret the linguistic styles that are more reflective of a younger demographic and present a more expressive language like the gender-ambiguous category. Additionally, compared to the fine-tuned models, LLMs can show a similar drop in QA performance in a zero-shot setting. However, in 2-shot, we see a significant improvement compared to fine-tuned models. This is a notable finding especially because it suggests that providing LLMs with even a minimal amount of context-specific examples can dramatically improve their performance. 

We make the following contributions in this paper: (1) We use LLMs for their text-generation capabilities to generate socio-demographically conditioned paraphrases, identifying which prompting structures are efficient based on the alignment (agreement) of the gender-oriented paraphrases with non-LLM methods;
(2) We conduct robustness evalutions for LLMs in zero- and two-shot settings and contrast their performance with fine-tuned models;
(3) We perform human evaluations of the paraphrases to investigate the correlation between QA misclassifications and paraphrase nuances;
(4) We conduct fine-grained reliability tests to examine the model performance across paraphrase properties such as similarity, explainability, and atomic questions.

\section{Related Work}
The robustness of language models to paraphrasing has been a significant area of study, with several key findings. \citet{shi2019robustness} highlights the robustness issues in paraphrase identification models when facing modifications with shared words. They demonstrate that these modifications can lead to a dramatic drop in model performance, and propose adversarial training as a potential solution. 
\citet{alting2020evaluating} extend this work to question answering models, showing that paraphrased questions can decrease their performance. \citet{wang2020cat} introduced the Controlled Adversarial Text Generation (CAT-Gen) model, which generates diverse and fluent adversarial texts to improve model performance. We have also seen that simple variations in prompts can greatly affect the model confidence in the same discussion \cite{khatun-brown-2023-reliability}.

Early studies, such as those by \citet{argamon2003gender}, laid the foundations for linguistic attributes by demonstrating clear differences in language usage between males and females. They found that females are more likely to use pronouns, whereas males frequently use noun specifiers, highlighting an "involved" versus "informational" style in communication. Building on this, \citet{newman2008gender} confirmed that not only do men and women use language differently, but these differences are consistent across different forms of communication, including social media posts and product reviews. They also suggested that these linguistic markers could predict the gender of the author with significant accuracy. \citet{mohtasseb2010affects} further explored this, finding that male writing style may be more consistent than female style, and that age can also influence writing style. 
We explore a new approach for measuring robustness where we create demographic sets meant to adversarially test the models' performance across different linguistic variations. We use the capabilities of the LLM itself as an agent that regenerates a given text in the requested demographic style. Here we evaluate which prompt engineering techniques are useful to obtain consistently varied paraphrases and how well the generated paraphrases align with real-world text. We also demonstrate that with minimal data augmentation, language models perform better on out-of-domain stylized text in QA tasks.

\section{Methodology}

\begin{figure*}[t]
    \centering
    \includegraphics[scale=0.80]{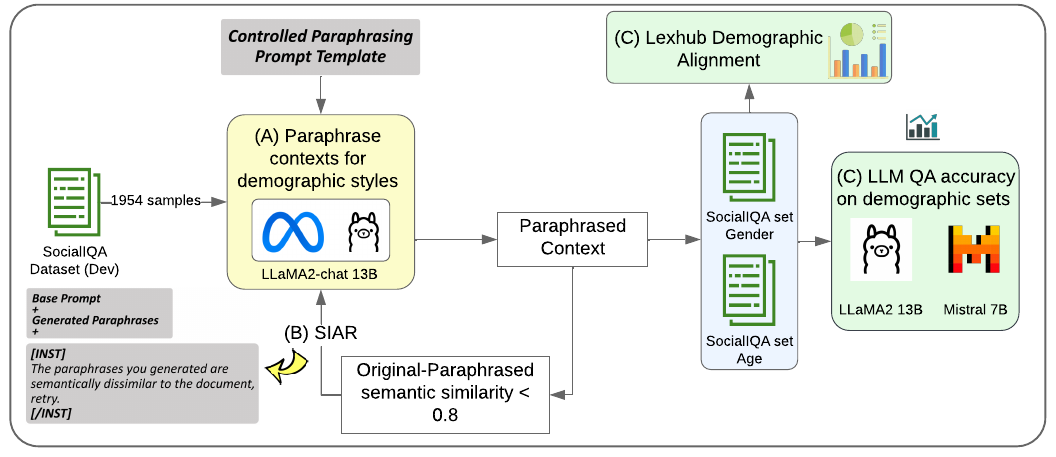}
    \caption{The workflow of our framework.}
    \label{fig:robustness_workflow}
\end{figure*}

Our evaluation framework consists of several components that can be seen in Figure \ref{fig:robustness_workflow}. Overall, the process consists of a series of steps to analyze the robustness to paraphrased sets using the SocialIQA (SIQA) dataset. In part A, in Figure \ref{fig:robustness_workflow}, we start by using the \texttt{LLAMA2} model to create paraphrases of the SIQA dataset's validation set, aiming to mirror different demographic linguistic styles without changing the meaning of the original context.  Next, in part B, we make sure that the paraphrases with low semantic similarity undergo refinement through semantic information augmented regeneration of paraphrases to ensure they preserve the original intent of the context.  In C, the alignment of these paraphrases with their intended demographic characteristics is verified using LexHub's demographic-based lexica. Finally, we test the LLMs' performance in question answering based on the stylized paraphrases, determining how paraphrasing affects the models' comprehension. 

\subsection{Experimental Setup}

\noindent
We employ LLMs, specifically \texttt{LLAMA2-chat-13B} \cite{touvron2023llama}, which will be referred to as \texttt{LLAMA2} hereafter, for paraphrasing prompts in distinct linguistic styles for the SIQA dataset as our primary corpus. Each sample from the validation set, with 1954 samples is structured in a prompt, question, and choices format, and each is independently paraphrased using the \texttt{LLAMA2} model. Furthermore, for robustness evaluation on the paraphrased sets we consider two models \texttt{LLAMA2} and \texttt{MISTRAL-instruct-7B} \cite{jiang2023mistral} (referred to as \texttt{MISTRAL} hereafter). Since inference is our primary goal we use Python bindings for the lightweight llama.cpp implementation which enables LLM inference with minimal setup. For more details refer to Table \ref{tab:hyperparameters} in the appendix.

\noindent
\paragraph{Dataset.} We construct our paraphrased QA sets using the SIQA \cite{sap2019socialiqa} dataset which comprises 38,000 multiple-choice questions focused on commonsense reasoning in social contexts. The dataset covers nine types of reasoning, such as intent, need, reaction, and effect. 
For our experiments, we have used the validation set with 1954 samples to get comparable benchmarks with the BIG-Bench \cite{ghazal2013bigbench} baselines. 

\noindent

\subsection{Paraphrase Generation}
We test different prompting styles to make the language model paraphrase the given text in a particular style:

\noindent
\paragraph{Controlled Paraphrasing.} The prompt consists of a basic task introduction the <<sys>>\{\}<</sys>> block, and a detailed instruction setup broader [INST]\{\}[/INST] block. We set up the \texttt{LLAMA2} model for the paraphrasing task by providing the system introduction \textit{"You are an English language Expert..."}. Once the system is defined, we provide the task information with detailed paraphrasing instructions in the INST block. Here we specify what the model precisely needs to do along with a set of rules which prevent the model from deviating from the given output format. Using this set of detailed instructions, we control the production of consistent paraphrases.

\begin{figure}[t]
\centering
\includegraphics[scale=0.4]{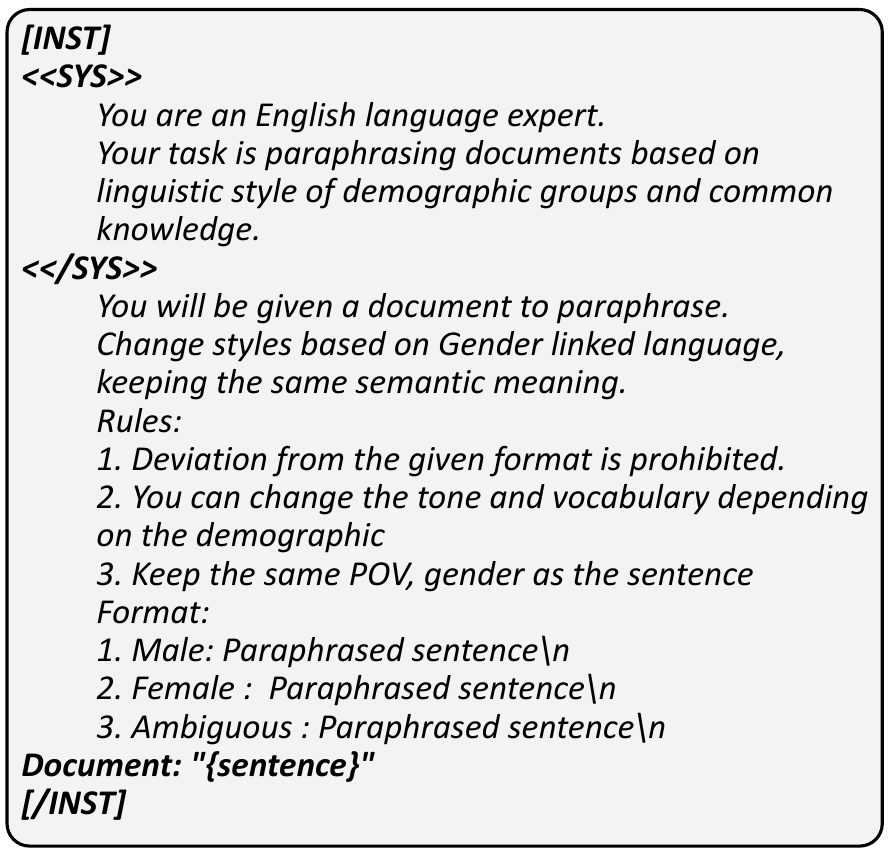}
\caption{Controlled paraphrasing. POV in the prompt means point of view.}
\label{fig:baseprompt}
\end{figure}
 
In this approach, \texttt{LLAMA2} is prompted with less specific guidance for the demographic style itself. This allows the model to autonomously infer the style characteristics based on the demographic category provided. This method leverages \texttt{LLAMA2}'s extensive pre-trained knowledge base and its ability to effectively contextualize input data by presenting only the demographic category (e.g., Male, Female, Young, Old). The base prompt (Figure \ref{fig:baseprompt}) shows how we instruct the model to directly perform paraphrasing based on its understanding of the demographic styles.

\begin{figure}[t]
\includegraphics[scale=0.46]{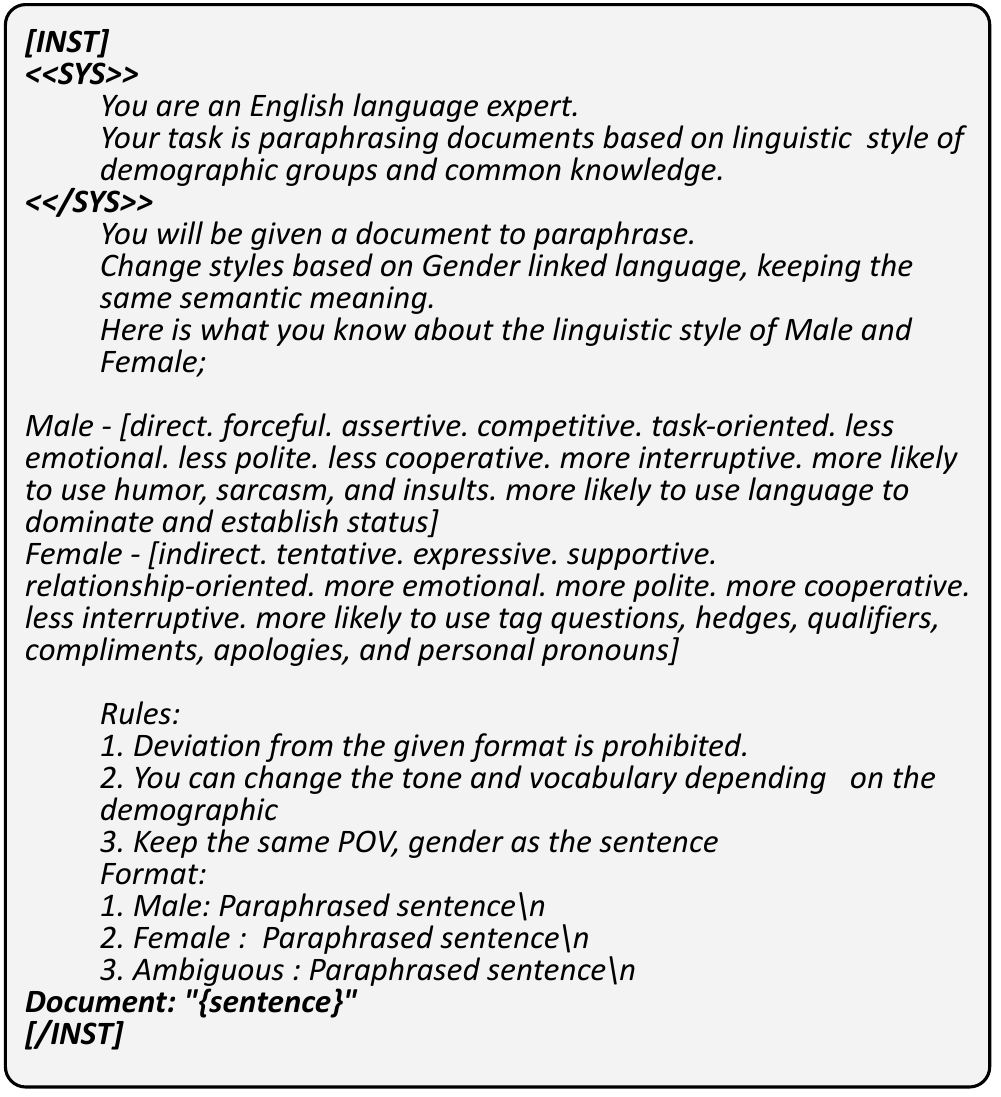}
\caption{Style-guided paraphrasing. POV in the prompt means point of view.}
\label{fig:style_guided}
\end{figure}

\noindent
\paragraph{Style-guided Paraphrasing.} In this method, which is an extension of the controlled paraphrasing prompt, we provide the model with detailed, explicit descriptions of the desired linguistic style traits associated with each demographic group. To guide \texttt{LLAMA2} in producing paraphrases reflective of specific linguistic styles, we explore common styles associated with the given demographic group for which the paraphrasing is requested.  
Works such as \citet{piersoul2023men} and \citet{preotiuc2016discovering} have captured distinctions in linguistic styles across different user attributes: gender, age, and occupation. Using this information, we experiment with delineating distinct linguistic attributes; for instance, for males and females\footnote{Considering binary genders in this work is only one perspective and is used to highlight the differences with a sample pair. Future work can investigate how this varies for other genders.}, emphasizing differences in directness, emotional expression, politeness, and syntactic complexity. This approach is grounded in empirical social media text analyses and linguistic studies that outline specific stylistic tendencies \cite{vikatos2017linguistic, newman2008gender}. 
We provide this information about the male and female style characteristics in Figure \ref{fig:style_guided}, with the additional instruction \textit{"Here is what you know about the linguistic style of Male and Female..."}. \\
By establishing these two different methods of prompting, controlled and style-guided, we examine which method provides paraphrases that are more aligned with the actual paraphrasing style of the given groups. This is quantified with the help of demographic-based lexica which we describe in the following section.\\

\subsection{Paraphrase Assessment}
After generating the paraphrases, we assess them regarding stylistic features and correlation with perplexity:

\paragraph{Stylistic Features.} LexHub is an online demographic prediction tool\footnote{\url{http://lexhub.org/wlt/lexica.html}} which is presented as a practical application of demographic-based lexica. By leveraging LexHub, we evaluate \texttt{LLAMA2} generated paraphrases for agreement with the lexica, ensuring alignment with demographic characteristics. \citet{sap2014developing} describe the creation of predictive lexica for age and gender using regression and classification models from social media data. LexHub achieved state-of-the-art accuracy in predicting age and gender over user-generated text on Facebook and Twitter. 

For Age, the model directly outputs representative age for the given sample, while for Gender the score is a relative measure, where >0 implies Female and <0 implies Male.  
We measure the model alignment using \textit{paraphrase agreement percentage} with the LexHub scores. First, we calculate the count of agreeing paraphrases; for Gender alignment, the following condition should hold: 
\[N(\texttt{lex\_score}(Par_F) - \texttt{lex\_score}(Par_M) > 0)\] 

\noindent
Similarly, for Age alignment, the count can be calculated by:  
\[
N(\texttt{lex\_score}(Par_O) - \texttt{lex\_score}(Par_Y) > 0)
\]

Using the above formulations, we calculate the count of paraphrase pairs where Female style paraphrase (\(Par_F\)) scores higher than the Male style paraphrase (\(Par_M\)) and Old style (\(Par_O\)) has a predicted age higher than Young style paraphrase (\(Par_Y\)). For alignment, we calculate the percentage of paraphrase pairs in Gender and Age demographic sets where the above conditions hold. It is calculated by dividing the count of agreeing pairs by the total number of samples that are paraphrased. We use 10\% of randomly sampled contexts from the SIQA validation set to measure the alignment. 

\begin{table}[t]
\centering
\small
\begin{tabularx}{\columnwidth}{lcXXr}
\toprule
Prompt style & \makecell[l]{LexHub\\Alignment} & \multicolumn{3}{c}{Cumulative scores} \\ 
&& Ambig & Female & Male \\
\midrule
Controlled&72\%& 1.33& 235.90 & -248.36  \\ 
\midrule
Style-guided&64\% & 12.49& 112.67 & -86.97  \\ 
\bottomrule
\end{tabularx}
\caption{Paraphrase alignment with LexHub for controlled vs. style-guided paraphrasing with \texttt{LLAMA2}. The cumulative scores show the raw aggregated scores for the paraphrased samples. Ambig is short for ambiguous.}
\label{table:paraphrase_alignment}
\end{table}

Using alignment as a metric to quantify the paraphrase quality, we examine which method of prompting produces better-aligned paraphrases, controlled or style-guided. Upon evaluation, we found that the alignment of \texttt{LLAMA2} with LexHub using controlled paraphrasing for Gender paraphrasing was higher (72\%) than the style-guided paraphrasing (64\%) (Table \ref{table:paraphrase_alignment}). 
This highlights that explicitly providing the style characteristics of the groups does not necessarily help the model produce better-aligned paraphrases. Following this conclusion, we conduct our evaluation study with the paraphrased sets generated with the controlled paraphrasing prompt.

\begin{figure}[t]
\centering
\includegraphics[scale=0.15]{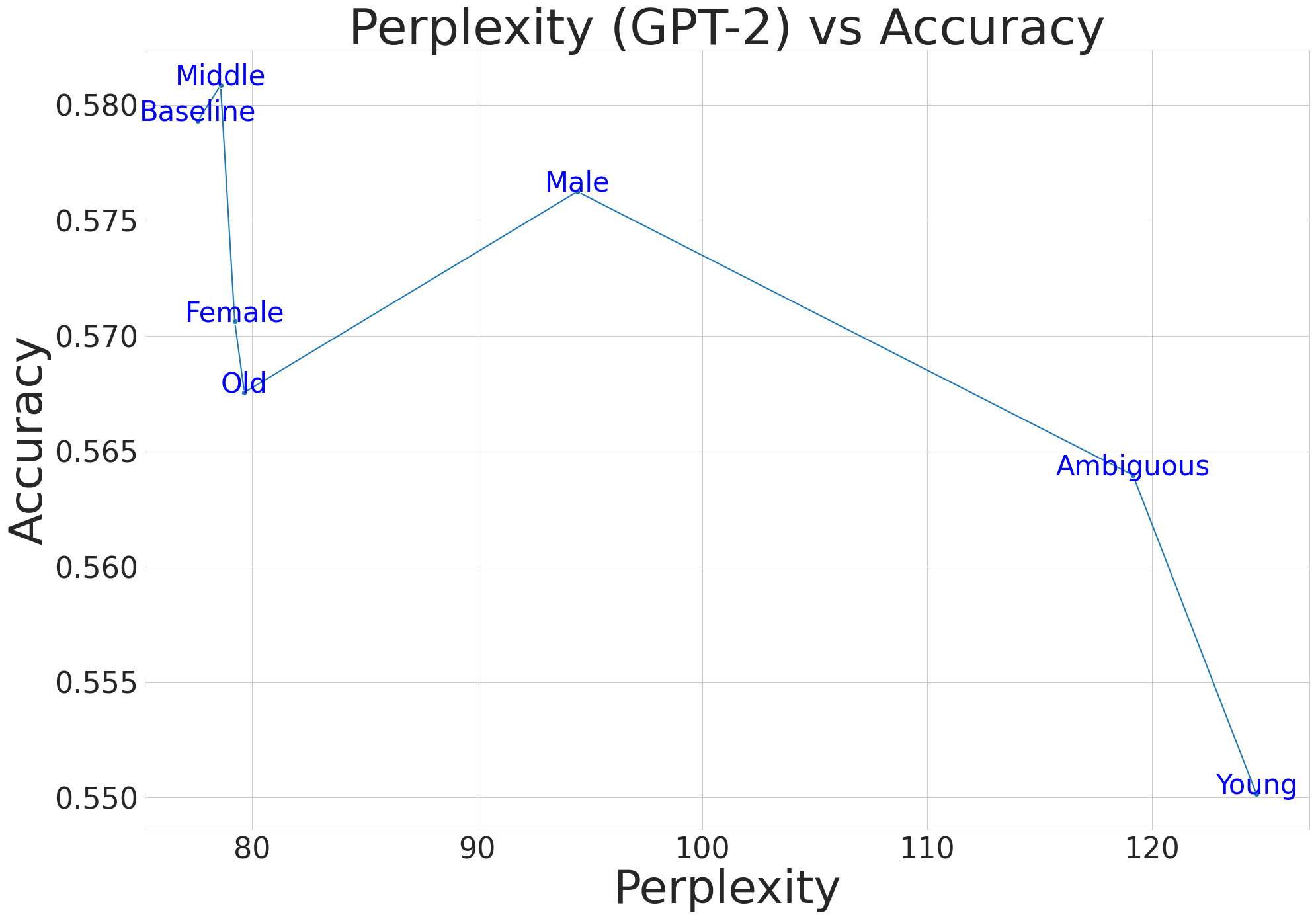}
\caption{Model performance (QA accuracy) vs. perplexity of the demographic set}
\label{fig:perplexity}
\end{figure}

\paragraph{Correlation with Perplexity.} Perplexity is a measure of a model's uncertainty in predicting the next word in a sequence. In the context of linguistic variation, given that the content of the sentence is the same, perplexity can also be interpreted as how comfortable the model is with predicting that style. The perplexity scores reported in Table \ref{table:results} for \texttt{LLAMA2} and GPT-2 across different demographic categories reveal significant insights into model behavior. Notably, we see an inverse relationship between perplexity and model performance; as perplexity increases, we observe a corresponding decrease in model accuracy. Figure \ref{fig:perplexity} shows this correlation. We consider GPT-2 as an \textit{unbiased third model} to calculate the set perplexity because the language is generated by the LLM itself. For GPT-2, results with the 0-shot setup demonstrate an inverse correlation between perplexity and model accuracy and higher perplexity scores directly translate to lower accuracy. This is also evident, albeit not so strongly, in the 2-shot setting. This possibly indicates that when the model is allowed to learn the language nuances in a low data setup, language complexity (perplexity) plays a smaller role.

\begin{table*}[t]
    \centering
    \small
    \begin{tabular}{l|c|c|c|c|c|c}
        \toprule
        &\multicolumn{2}{c|}{\texttt{LLAMA2}} &\multicolumn{2}{c|}{\texttt{MISTRAL}}&\multicolumn{2}{c}{Perplexity} \\
        \textbf{QA Set} & \textbf{(0-shot)} & \textbf{(2-shot)} & \textbf{(0-shot)} & \textbf{(2-shot)} & \textbf{\texttt{LLAMA2}} & \textbf{GPT-2} \\
        \midrule
        Original & 57.88 & 62.42 & 66.37 & 71.44 & 46.940 & 77.58 \\
        Male & 57.58 & 63.15 & 64.27 & 67.50 & 27.183 & 94.45 \\
        Female & 56.80 & 63.66 & \textbf{63.76} &  69.03 & 24.558 & 79.22 \\
        Ambiguous & 56.19 & \textbf{61.66} & 64.38 & \textbf{66.78} & 33.862 & 119.16 \\
        Young (<20) & \textbf{54.63} &  61.77 & 63.97 & 67.09 & 26.976 & 124.65 \\
        Middle Age (20-60) & 57.72 & 64.68 & 65.50 & 68.88 & 18.845 & 78.59 \\
        Old (>60) & 56.75 & 63.05 & 65.45 & 68.16 & 20.626 & 79.64 \\
        \bottomrule
        \hline
    \end{tabular}
    \caption{Performance on paraphrased sets. Accuracy is reported in percentage.}
    \label{table:results}
\end{table*}

\section{Robustness Evaluation}
Table \ref{table:results} shows the performance of LLMs of our demographically paraphrased SIQA test set. The performance of 0-shot and 2-shot inference on demographic sets is compared against the baseline dataset for the models \texttt{LLAMA2} and \texttt{MISTRAL}.

\subsection{Decoder Models}

\begin{table}[t]
\small
\centering
\setlength{\tabcolsep}{4pt}
\begin{tabular}{l|c|c}
\toprule
 &\multicolumn{2}{c}{\textbf{Fine-tuned Models}} \\
\textbf{QA Set} & \textbf{BERT-base-uncased} & \textbf{RoBERTa-base}  \\
\midrule
Original & 56.03 & 65.76 \\
Male & 55.24 (-1.41\%) & 63.40 (-3.59\%) \\
Female & 55.37 (-1.18\%) & 63.56 (-3.34\%) \\
Ambiguous & 54.75 (-2.28\%) & 62.48 \textbf{(-4.99\%)} \\
Young (<20) & 54.04 \textbf{(-3.55\%)} & 63.05 (-4.12\%) \\
Middle (20-60) & 55.01 (-1.82\%) & 63.61 (-3.27\%) \\
Old (>60) & 55.11 (-1.64\%) & 63.30 (-3.74\%) \\
\bottomrule
\end{tabular}
\caption{Performance of encoder models (BERT and RoBERTa) on original and paraphrased SIQA sets}
\label{table:siqa_finetuned}
\end{table}

The results summarized in Table \ref{table:results} highlight the performance discrepancies across different demographic groups. The evaluation metrics are done on both 0-shot and 2-shot inference approaches, providing insight into the models' ability to generalize from minimal examples. For the original test set, the \texttt{LLAMA2} model demonstrates a 0-shot accuracy of 57.88\%, improving to 62.42\% with 2-shot inference. \texttt{MISTRAL} expectedly shows a superior baseline performance at 66.37\% (0-shot), which improves to 71.44\% (2-shot). This pattern of improvement with additional shots is consistent across demographic categories for both models, indicating that even minimal set of examples helps in improving LLM performance.

Demographic analysis reveals nuanced performance variations. For instance, in the Male and Female categories, the models show a slight decrease in 0-shot accuracy compared to the original set, yet they improve in 2-shot, particularly notable in the Female category with \texttt{LLAMA2} achieving a 63.66\% (2-shot) accuracy. \texttt{MISTRAL}, on the other hand, turns out to be less robust with the female style in the 0-shot setting while showing the highest performance in the 2-shot setting with the same set. Ambiguous gender representation poses a more significant challenge, with a marked reduction in performance in both 0-shot and 2-shot inference.

Age-based disparities are evident as well. The Young demographic witnesses the most significant drop in performance for \texttt{LLAMA2} in the 0-shot scenario at 54.63\%. This is seen in the 2-shot setting as well, where the Young demographic shows a lower accuracy. Conversely, the Middle Age and Old groups showcase more resilient performance metrics, aligning closely with the original dataset outcomes. This resilience is particularly pronounced in the Middle Age group, where \texttt{LLAMA2} achieves its highest 2-shot accuracy at 64.68\%. \\
\noindent
\subsection{Encoder Models} In this analysis, we explore the effects of demographic-specific paraphrasing on the accuracy of two encoder models, BERT-base-uncased \cite{devlin2018bert} and RoBERTa-base \cite{liu2019roberta}. Before evaluation, both models were fine-tuned on the SIQA training set for 10 epochs. Following fine-tuning, the performance of both models on various demographic paraphrases was assessed. In Table \ref{table:siqa_finetuned}, we see that BERT-base-uncased started with a baseline accuracy of 56.03\% on the original set. When applied to demographically paraphrased sets, there was a noticeable decline across all demographics, with the most significant drop exhibited in the Young demographic at 3.55\%. The RoBERTa-base model, which began with a higher baseline accuracy of 65.76\% on the original set, also showed decreased performance across the paraphrased sets, with the Ambiguous category facing the steepest decline at 4.99\%.

Comparing these results to those of the decoder models, \texttt{LLAMA2} and \texttt{MISTRAL}, a similar trend is revealed where both sets of models experience performance drops when evaluated with demographically paraphrased questions. This suggests that despite the sophistication of LLMs and better generalizing capacity, there are inherent challenges in addressing demographic-specific language that transcends the particularities of individual model architectures. 

\begin{table*}[t]
    \centering
    \small
    \begin{tabular}{l|c|c|c|c|c|c}
        \toprule
        & \multicolumn{3}{c|}{\texttt{LLAMA2}} & \multicolumn{3}{c}{\texttt{MISTRAL}} \\
        \midrule
        \textbf{QA Set} & \textbf{(0-shot)} & \textbf{(1 Regen)} & \textbf{(3 Regen)} & \textbf{(0-shot)} & \textbf{(1 Regen)}  & \textbf{(3 Regen)}  \\
        \midrule
        Male & 57.58 & 57.57 (-0.01\%) & 58.13 (+0.96\%) & 64.27 & 64.12 (-0.23\%) & 64.43 (+0.25\%) \\
        Female & 56.80 & 57.47 (+1.18\%) & 57.76 (+1.69\%) & 63.76 & 63.51 (-0.39\%) & 64.32 (+0.88\%) \\
        Ambiguous & 56.19 & 56.96 (+1.37\%) & 56.90 (+1.26\%) & 64.38 & 64.68 (+0.46\%) & 64.79 (+0.64\%) \\
        Young  & 54.63 & 54.29 (-0.62\%) & 54.24 (-0.71\%) & 63.97 & 63.81 (-0.25\%) & 64.43 (+0.72\%) \\
        Middle Age & 57.72 & 57.83 (+0.19\%) & 57.80 (+0.14\%) & 65.50 & 65.86 (+0.54\%) & 65.80 (+0.46\%) \\
        Old & 56.75 & 56.29 (-0.81\%) & 55.98 (-1.36\%) & 65.45 & 65.50 (+0.07\%) & 65.64 (+0.29\%)  \\
        \bottomrule
    \end{tabular}
    \caption{Performance comparison of paraphrases generated with Semantic Information Augmented Re-generation using embedding similarity of original paraphrases as the augmented information in the prompt (0-shot QA evaluation)}
    \label{table:results_regen}
\end{table*}

\begin{figure}[t]
\centering
\includegraphics[scale=0.14]{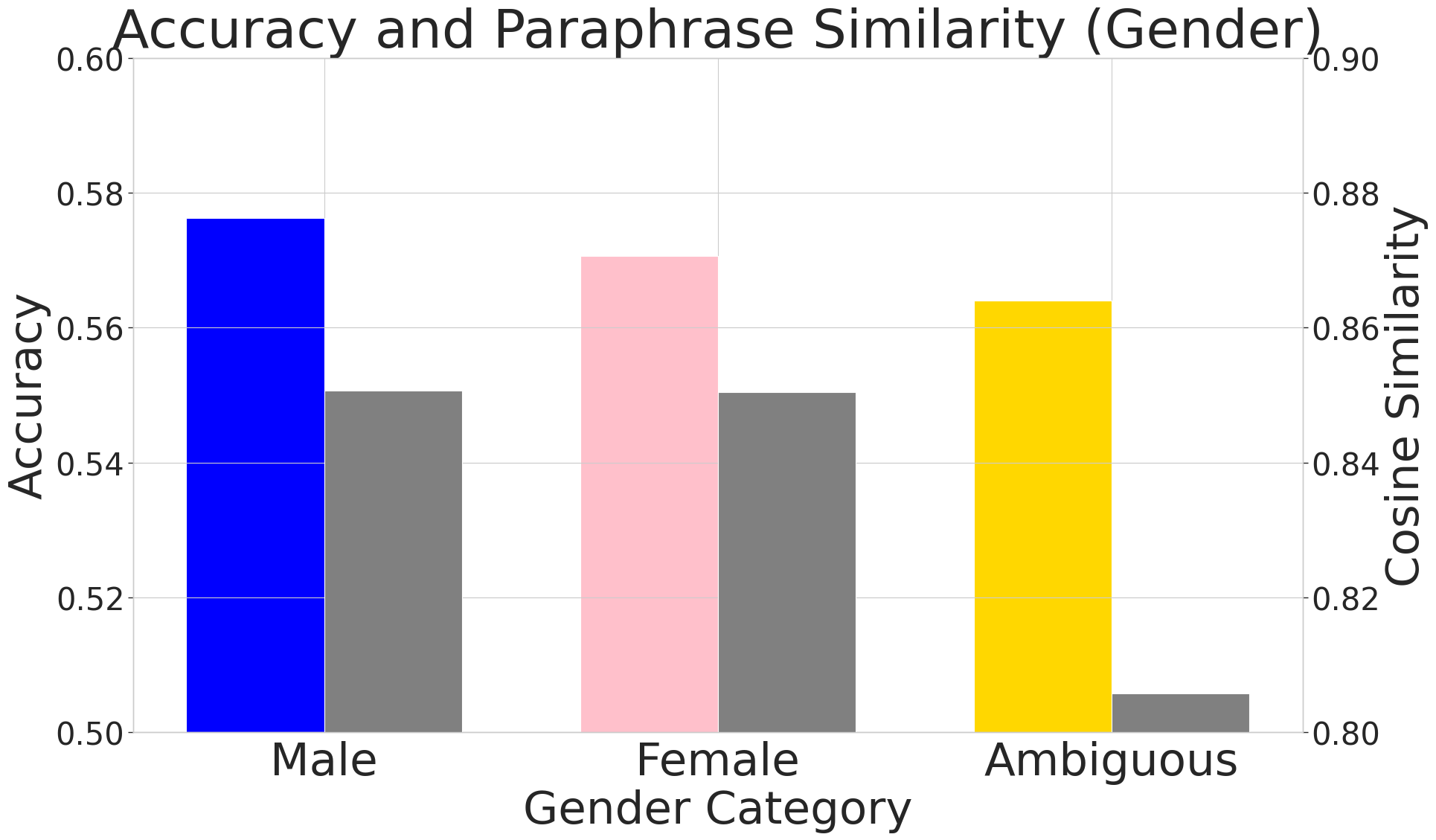}

\includegraphics[scale=0.14]{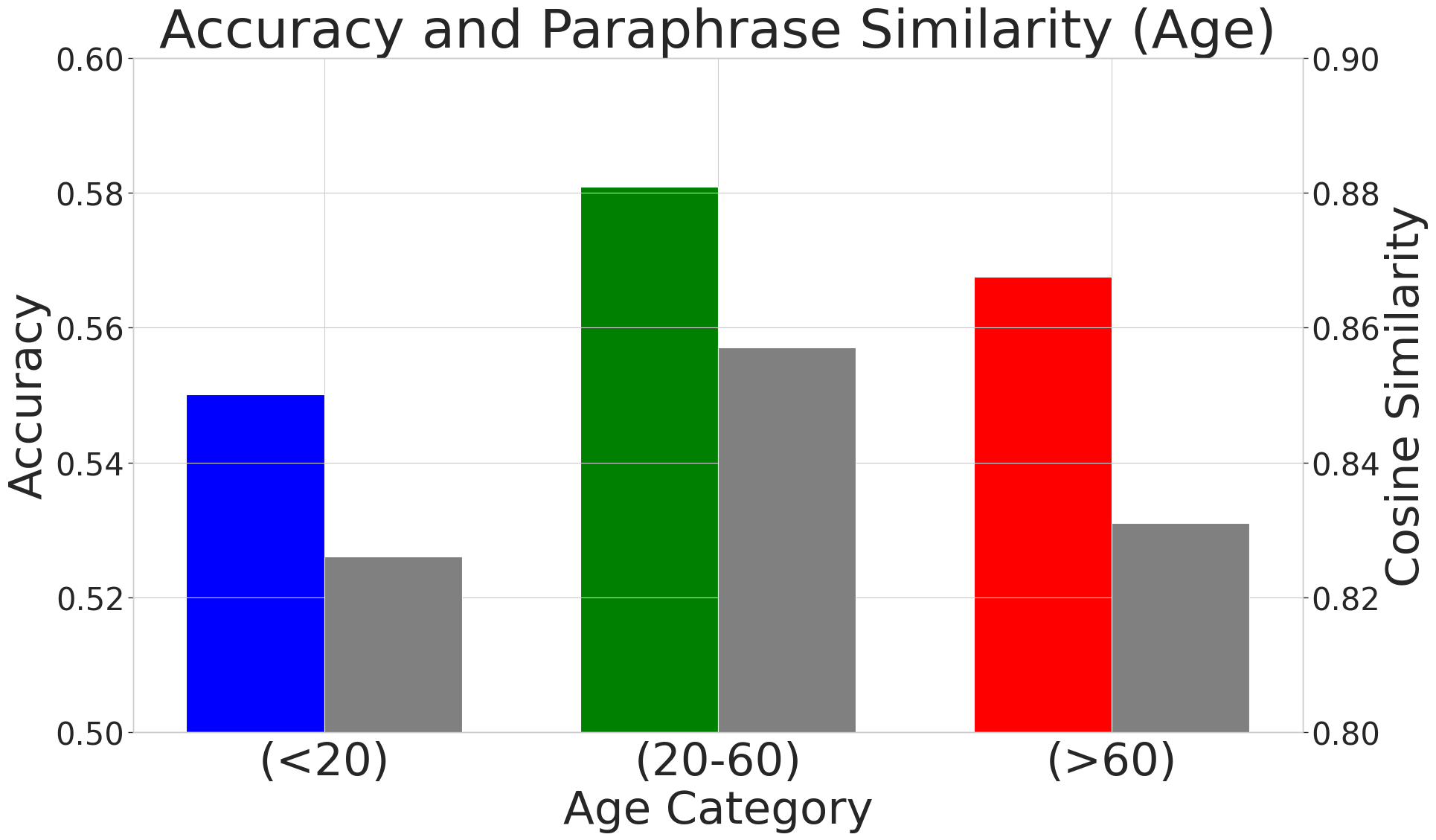}
\caption{Performance with paraphrased sets contrasted with the semantic similarity of the paraphrased set with the SIQA validation set. The colored bars represent the demographic set accuracy while the gray bars show their respective cosine similarity score.}
\label{fig:gen_age_qa_accuracy}
\end{figure}

\noindent
\subsection{Semantically Augmented Re-generation} 
When we examine the semantic closeness of the paraphrased sets to the original contexts (Figure \ref{fig:gen_age_qa_accuracy}), we observe that the Ambiguous gender category and the Young age group are the lower-performing sets in terms of both prediction accuracy and average semantic similarity scores when compared to their counterparts and the original sets. This highlights a correlation between the quality of the generated paraphrases for these sets and the performance on the QA task. Table \ref{table:results_regen} presents a detailed performance comparison of paraphrases generated using semantic information augmented re-generation, which incorporates embedding similarity of paraphrased sentences to the original context as the augmented information in the prompt. We perform this assessment on zero-shot QA evaluation. The embeddings are derived using the Sentence Transformers architecture with the pre-trained weights of \textit{paraphrase-distilroberta-base-v2} \cite{reimers-2019-sentence-bert}. The model has been trained on a large corpus of paraphrase data, making it suitable for tasks involving paraphrase identification. 

For the \texttt{LLAMA2} model, results indicated nuanced differences across demographic groups when comparing the effects of one regeneration versus three. For instance, the Male group saw an improvement increase from a negligible decrease after one regeneration to a 0.96\% improvement after three regenerations. The Female group demonstrated a similar progressive improvement, from 1.18\% after one regeneration to 1.69\% after three, showcasing a clear benefit of multiple regeneration cycles in refining paraphrases. In contrast, the Young group's performance continued to decline, with -0.62\% after one regeneration and -0.71\% after three, indicating that semantic reinforcement may not be enough to enhance understanding of the Young style.

The \texttt{MISTRAL} model displayed somewhat different dynamics. The Male group initially experienced a decrease of 0.23\% after one regeneration but improved by 0.25\% after three regenerations. The Female group also showed significant improvement over time, with a slight initial drop turning into a 0.88\% gain after three regenerations. This pattern was also visible albeit less pronounced in the Old group which already shows a high 0-shot accuracy. We see slight but steady improvements over the initial performance, highlighting a slow but beneficial adaptation to regeneration.

When comparing the two models, \texttt{LLAMA2} tends to show more significant performance fluctuations across multiple regenerations, indicating a potential for higher gains but also highlighting its sensitivity to the regeneration process. Meanwhile, \texttt{MISTRAL} tends to show a steadier progression or recovery in performance, especially notable in groups that initially experience a drop. Overall, the analysis shows that generally, multiple regenerations can enhance paraphrase quality.

\section{Analysis of Reliability Factors} 

\subsection{Human Evaluation of Paraphrases}
We conducted a detailed analysis of paraphrase predictions to assess the impact of stylistic and semantic changes on their validity across male and female sets. The misclassifications were categorized into two primary types: incorrect paraphrases and hallucinations. Incorrect paraphrases were those that did not preserve the original meaning, while hallucinations involved the addition of information not present in the source text. 

We analyzed the misclassification that are common (shared between male and female sets), male-only, and female-only (Table \ref{tab:paraphrase_validity}). The common set, which encompasses errors found in both male and female samples, shows an invalid paraphrasing rate of approximately 25\%, including both incorrect renditions and hallucinations. This rate is higher than that observed in the male-only set but lower than in the female-only set. Such a pattern suggests a slight correlation between the overall misclassification rates for the male and female sets, possibly hinting that the model adds more artifacts to the paraphrases for the female context. Generally, these artifacts are related to verbs of niceness/politeness which alter the meaning of the original sentence.

Despite these discrepancies in misclassification rates, it is important to note that the majority of paraphrases across all sets are rendered correctly. Over 75\% of paraphrases in the common set, nearly 87\% in the male-only set, and about 64\% in the female-only set are valid. This implies that the paraphrasing errors are more often due to subtle nuances in language usage that the model misinterprets, rather than a direct failure to capture and convey the original semantics. 

\begin{table}
\centering
\small
\begin{tabular}{l|c|c|c}
\toprule
\textbf{Set} & \textbf{Incorrect} & \textbf{Hallucinations} & \textbf{Valid} \\ \midrule
Common       & 15.28 \%                   & 9.72 \%                        & 75.00 \%               \\ \hline
Male         & 4.35 \%                    & 8.70 \%                         & 86.96 \%               \\ \hline
Female       & 3.57 \%                    & 32.14 \%                        & 64.29 \%               \\ \bottomrule
\end{tabular}
\caption{Percentage of Invalid Paraphrasing in Misclassified Samples}
\label{tab:paraphrase_validity}
\end{table}

\subsection{Effects of Language Diversity}
The exploration of language diversity's impact on QA performance uncovers notable trends in model robustness across various demographic sets. Specifically, our analysis indicates that the Young and Gender Ambiguous demographic sets exhibit the least robustness in QA tasks when compared to the rest of the categories, such as Male, Female, Middle Age, and Old (Figure \ref{table:results}). 
Furthermore, the reduction in performance seems to stem from the nature of the language employed in the paraphrasing for these demographics, which tends to be more expressive, with a higher perplexity, and less formal compared to other groups (Table \ref{table:distinct_words} in Appendix).  It suggests that the formality of the language may significantly influence the model's ability to process information. This implies that the commonsense reasoning capacity of a model could vary with the formal structure of the language, with less formally written language presenting a greater challenge for accurate comprehension of the social or commonsense context. Such findings highlight a crucial consideration for developing more robust QA models, taking into account the linguistic diversity inherent in human language such as informality or expressiveness.

\subsection{Semantic Differences in Paraphrasing}
\begin{figure}[t]
\centering
\includegraphics[scale=0.14]{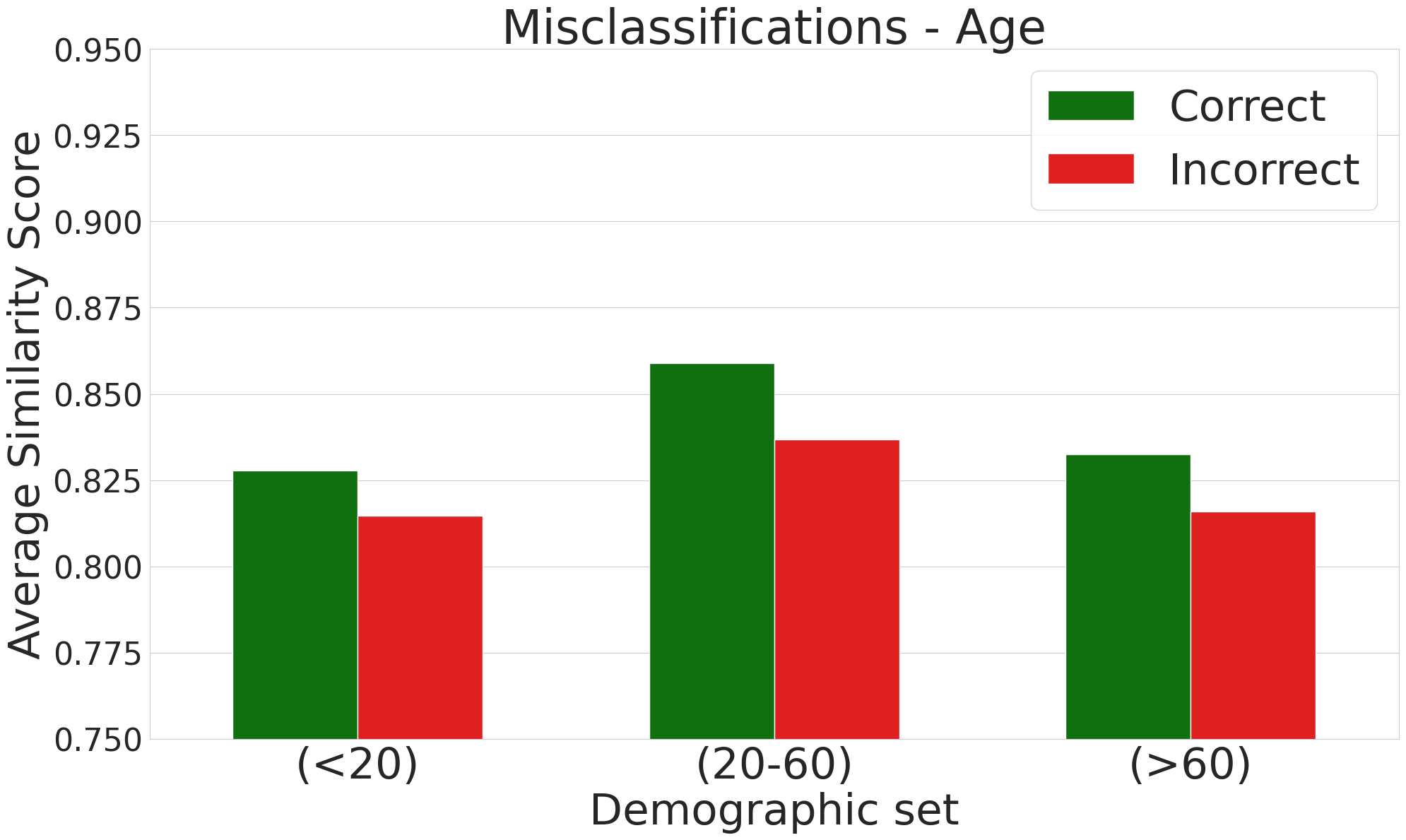}

\includegraphics[scale=0.14]{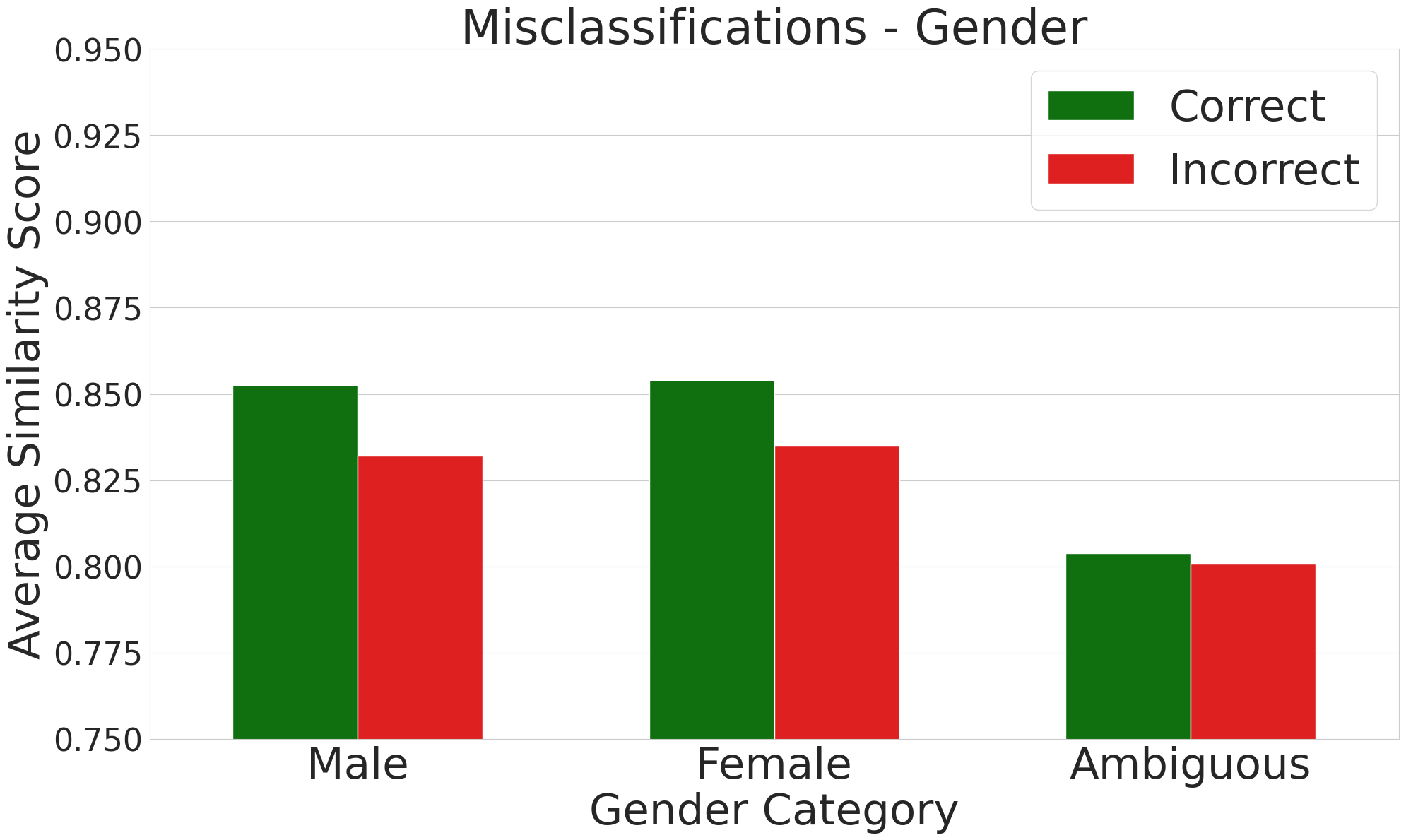}
\caption{Comparison of average semantic similarity for paraphrase-original pairs for correctly classified (green) vs misclassified samples (red). Correctly classified samples consistently show higher average similarity than misclassified ones across demographic sets.}
\label{fig:semantic_sim}
\end{figure}

We investigate the role of semantic differences in paraphrases and their impact on misclassifications in QA tasks (Figure \ref{fig:semantic_sim}). We measure the cosine similarity between the original and paraphrase embeddings obtained from the \textit{paraphrase-distilroberta} model \cite{reimers-2019-sentence-bert}.  
Upon manual verification, it was observed that out of the common misclassified samples in the Male and Female set, only about 20\% could be accounted for by incorrect/hallucinated paraphrasing. The rest of the misclassifications were likely due to linguistic changes.

\section{Conclusion}
Our analysis revealed that demographic nuances alter the model's ability to accurately interpret and answer questions. Paraphrases tailored to younger demographics, which often use informal expressions, tend to decrease model accuracy. This suggests a potential gap in the model's training data or its ability to handle informal language. On the other hand, paraphrases associated with Old, Male, and Middle Age demographics, which typically feature more formal language, showed better compatibility with the model’s predictive capabilities, indicating that LLMs are more familiar with less expressive language and a more formal tone. Our findings underscore the need for more robust models across the linguistic styles of different demographics.

\section{Limitations and Ethical Considerations}
The approach that we utilized in this paper, of rephrasing language into sociodemographic styles presents some limitations. Primarily, the generalizability of our findings might be limited as they are based on specific models and a dataset that may not fully capture the diverse nuances of language influenced by broader demographic factors such as ethnicity, regional dialects, or socio-economic status. 
We also acknowledge the limitations of using self-reported user categorization of text as the basis for strict classifications of gender and age. Hence, the categorizations (Male, Female, Young, Middle, and Old) we have used do not imply a strict distinction in categories but rather a relative orientation of the language which is based on the frequency of certain words in self-reported age and gender of text, which is what the LexHub is also based on.  
Our formula for calculating model alignment incorporates this relative orientation of language. We also try to improve the diversity of gendered language by incorporating a "Gender ambiguous" category to be more inclusive of language style non-conforming to the binary gender categories. 
Ethically, the incorporation of demographic-specific paraphrasing might raise concerns about reinforcing biases. Models that perform variably across different demographic groups could inadvertently perpetuate discrimination or unequal treatment. Therefore, it is crucial to approach the deployment of such technologies with a strong commitment to fairness, transparency, and inclusivity. 
                                                             
\section{Acknowledgements}

\bibliography{custom}

\begin{thebibliography}{26}
\expandafter\ifx\csname natexlab\endcsname\relax\def\natexlab#1{#1}\fi

\bibitem[{Alting~von Geusau and Bloem(2020)}]{alting2020evaluating}
Paulo Alting~von Geusau and Peter Bloem. 2020.
\newblock Evaluating the robustness of question-answering models to paraphrased questions.
\newblock In \emph{Benelux Conference on Artificial Intelligence}, pages 1--14. Springer.

\bibitem[{Argamon et~al.(2003)Argamon, Koppel, Fine, and Shimoni}]{argamon2003gender}
Shlomo Argamon, Moshe Koppel, Jonathan Fine, and Anat~Rachel Shimoni. 2003.
\newblock Gender, genre, and writing style in formal written texts.
\newblock \emph{Text \& talk}, 23(3):321--346.

\bibitem[{Bommasani et~al.(2023)Bommasani, Liang, and Lee}]{bommasani2023holistic}
Rishi Bommasani, Percy Liang, and Tony Lee. 2023.
\newblock Holistic evaluation of language models.
\newblock \emph{Annals of the New York Academy of Sciences}, 1525(1):140--146.

\bibitem[{Devlin et~al.(2018)Devlin, Chang, Lee, and Toutanova}]{devlin2018bert}
Jacob Devlin, Ming-Wei Chang, Kenton Lee, and Kristina Toutanova. 2018.
\newblock Bert: Pre-training of deep bidirectional transformers for language understanding.
\newblock \emph{arXiv preprint arXiv:1810.04805}.

\bibitem[{Ghazal et~al.(2013)Ghazal, Rabl, Hu, Raab, Poess, Crolotte, and Jacobsen}]{ghazal2013bigbench}
Ahmad Ghazal, Tilmann Rabl, Minqing Hu, Francois Raab, Meikel Poess, Alain Crolotte, and Hans-Arno Jacobsen. 2013.
\newblock Bigbench: Towards an industry standard benchmark for big data analytics.
\newblock In \emph{Proceedings of the 2013 ACM SIGMOD international conference on Management of data}, pages 1197--1208.

\bibitem[{Hutto and Gilbert(2014)}]{hutto2014vader}
Clayton Hutto and Eric Gilbert. 2014.
\newblock Vader: A parsimonious rule-based model for sentiment analysis of social media text.
\newblock In \emph{Proceedings of the international AAAI conference on web and social media}, volume~8, pages 216--225.

\bibitem[{Jiang et~al.(2023)Jiang, Sablayrolles, Mensch, Bamford, Chaplot, Casas, Bressand, Lengyel, Lample, Saulnier et~al.}]{jiang2023mistral}
Albert~Q Jiang, Alexandre Sablayrolles, Arthur Mensch, Chris Bamford, Devendra~Singh Chaplot, Diego de~las Casas, Florian Bressand, Gianna Lengyel, Guillaume Lample, Lucile Saulnier, et~al. 2023.
\newblock Mistral 7b.
\newblock \emph{arXiv preprint arXiv:2310.06825}.

\bibitem[{Khatun and Brown(2023)}]{khatun-brown-2023-reliability}
Aisha Khatun and Daniel Brown. 2023.
\newblock \href {https://doi.org/10.18653/v1/2023.trustnlp-1.8} {Reliability check: An analysis of {GPT}-3{'}s response to sensitive topics and prompt wording}.
\newblock In \emph{Proceedings of the 3rd Workshop on Trustworthy Natural Language Processing (TrustNLP 2023)}, pages 73--95, Toronto, Canada. Association for Computational Linguistics.

\bibitem[{Liu et~al.(2019)Liu, Ott, Goyal, Du, Joshi, Chen, Levy, Lewis, Zettlemoyer, and Stoyanov}]{liu2019roberta}
Yinhan Liu, Myle Ott, Naman Goyal, Jingfei Du, Mandar Joshi, Danqi Chen, Omer Levy, Mike Lewis, Luke Zettlemoyer, and Veselin Stoyanov. 2019.
\newblock Roberta: A robustly optimized bert pretraining approach.
\newblock \emph{arXiv preprint arXiv:1907.11692}.

\bibitem[{Marx~Larre(2023)}]{marx2023effects}
Miguel Marx~Larre. 2023.
\newblock Effects of paraphrasing and demographic metadata on nli classification performance.
\newblock {B.S.} thesis.

\bibitem[{Moeller et~al.(2023)Moeller, Nikolaev, and Pad{\'o}}]{moeller-etal-2023-attribution}
Lucas Moeller, Dmitry Nikolaev, and Sebastian Pad{\'o}. 2023.
\newblock \href {https://doi.org/10.18653/v1/2023.emnlp-main.980} {An attribution method for {S}iamese encoders}.
\newblock In \emph{Proceedings of the 2023 Conference on Empirical Methods in Natural Language Processing}, pages 15818--15827, Singapore. Association for Computational Linguistics.

\bibitem[{Mohtasseb and Ahmed(2010)}]{mohtasseb2010affects}
Haytham Mohtasseb and Amr Ahmed. 2010.
\newblock The affects of demographics differentiations on authorship identification.
\newblock \emph{Electronic Engineering and Computing Technology}, pages 409--417.

\bibitem[{Newman et~al.(2008)Newman, Groom, Handelman, and Pennebaker}]{newman2008gender}
Matthew~L Newman, Carla~J Groom, Lori~D Handelman, and James~W Pennebaker. 2008.
\newblock Gender differences in language use: An analysis of 14,000 text samples.
\newblock \emph{Discourse processes}, 45(3):211--236.

\bibitem[{Piersoul and Van~de Velde(2023)}]{piersoul2023men}
Jozefien Piersoul and Freek Van~de Velde. 2023.
\newblock Men use more complex language than women, but the difference has decreased over time: a study on 120 years of written dutch.
\newblock \emph{Linguistics}, 61(3):725--747.

\bibitem[{Preotiuc-Pietro et~al.(2016)Preotiuc-Pietro, Xu, and Ungar}]{preotiuc2016discovering}
Daniel Preotiuc-Pietro, Wei Xu, and Lyle Ungar. 2016.
\newblock Discovering user attribute stylistic differences via paraphrasing.
\newblock In \emph{Proceedings of the aaai conference on artificial intelligence}, volume~30.

\bibitem[{Reimers and Gurevych(2019)}]{reimers-2019-sentence-bert}
Nils Reimers and Iryna Gurevych. 2019.
\newblock \href {http://arxiv.org/abs/1908.10084} {Sentence-bert: Sentence embeddings using siamese bert-networks}.
\newblock In \emph{Proceedings of the 2019 Conference on Empirical Methods in Natural Language Processing}. Association for Computational Linguistics.

\bibitem[{Sanh et~al.(2019)Sanh, Debut, Chaumond, and Wolf}]{sanh2019distilbert}
Victor Sanh, Lysandre Debut, Julien Chaumond, and Thomas Wolf. 2019.
\newblock Distilbert, a distilled version of bert: smaller, faster, cheaper and lighter.
\newblock \emph{arXiv preprint arXiv:1910.01108}.

\bibitem[{Sap et~al.(2014)Sap, Park, Eichstaedt, Kern, Stillwell, Kosinski, Ungar, and Schwartz}]{sap2014developing}
Maarten Sap, Gregory Park, Johannes Eichstaedt, Margaret Kern, David Stillwell, Michal Kosinski, Lyle Ungar, and H~Andrew Schwartz. 2014.
\newblock Developing age and gender predictive lexica over social media.
\newblock In \emph{Proceedings of the 2014 conference on empirical methods in natural language processing (EMNLP)}, pages 1146--1151.

\bibitem[{Sap et~al.(2019)Sap, Rashkin, Chen, LeBras, and Choi}]{sap2019socialiqa}
Maarten Sap, Hannah Rashkin, Derek Chen, Ronan LeBras, and Yejin Choi. 2019.
\newblock Socialiqa: Commonsense reasoning about social interactions.
\newblock \emph{arXiv preprint arXiv:1904.09728}.

\bibitem[{Shi and Huang(2019)}]{shi2019robustness}
Zhouxing Shi and Minlie Huang. 2019.
\newblock Robustness to modification with shared words in paraphrase identification.
\newblock \emph{arXiv preprint arXiv:1909.02560}.

\bibitem[{Tan et~al.(2021)Tan, Joty, Baxter, Taeihagh, Bennett, and Kan}]{tan-etal-2021-reliability}
Samson Tan, Shafiq Joty, Kathy Baxter, Araz Taeihagh, Gregory~A. Bennett, and Min-Yen Kan. 2021.
\newblock \href {https://doi.org/10.18653/v1/2021.acl-long.321} {Reliability testing for natural language processing systems}.
\newblock In \emph{Proceedings of the 59th Annual Meeting of the Association for Computational Linguistics and the 11th International Joint Conference on Natural Language Processing (Volume 1: Long Papers)}, pages 4153--4169, Online. Association for Computational Linguistics.

\bibitem[{Tan et~al.(2023)Tan, Min, Li, Li, Hu, Chen, and Qi}]{tan2023evaluation}
Yiming Tan, Dehai Min, Yu~Li, Wenbo Li, Nan Hu, Yongrui Chen, and Guilin Qi. 2023.
\newblock Evaluation of chatgpt as a question answering system for answering complex questions.
\newblock \emph{arXiv preprint arXiv:2303.07992}.

\bibitem[{Touvron et~al.(2023)Touvron, Martin, Stone, Albert, Almahairi, Babaei, Bashlykov, Batra, Bhargava, Bhosale et~al.}]{touvron2023llama}
Hugo Touvron, Louis Martin, Kevin Stone, Peter Albert, Amjad Almahairi, Yasmine Babaei, Nikolay Bashlykov, Soumya Batra, Prajjwal Bhargava, Shruti Bhosale, et~al. 2023.
\newblock Llama 2: Open foundation and fine-tuned chat models.
\newblock \emph{arXiv preprint arXiv:2307.09288}.

\bibitem[{Vikatos et~al.(2017)Vikatos, Messias, Miranda, and Benevenuto}]{vikatos2017linguistic}
Pantelis Vikatos, Johnnatan Messias, Manoel Miranda, and Fabricio Benevenuto. 2017.
\newblock Linguistic diversities of demographic groups in twitter.
\newblock In \emph{Proceedings of the 28th ACM Conference on Hypertext and Social Media}, pages 275--284.

\bibitem[{Wang et~al.(2024)Wang, Morgenstern, and Dickerson}]{wang2024large}
Angelina Wang, Jamie Morgenstern, and John~P Dickerson. 2024.
\newblock Large language models cannot replace human participants because they cannot portray identity groups.
\newblock \emph{arXiv preprint arXiv:2402.01908}.

\bibitem[{Wang et~al.(2020)Wang, Wang, Qin, Packer, Li, Chen, Beutel, and Chi}]{wang2020cat}
Tianlu Wang, Xuezhi Wang, Yao Qin, Ben Packer, Kang Li, Jilin Chen, Alex Beutel, and Ed~Chi. 2020.
\newblock Cat-gen: Improving robustness in nlp models via controlled adversarial text generation.
\newblock \emph{arXiv preprint arXiv:2010.02338}.

\end{thebibliography}

\newpage
\appendix

\section{Explainability of Paraphrases with XSBERT. }
\begin{figure}[t]
\centering
\includegraphics[scale=0.17]{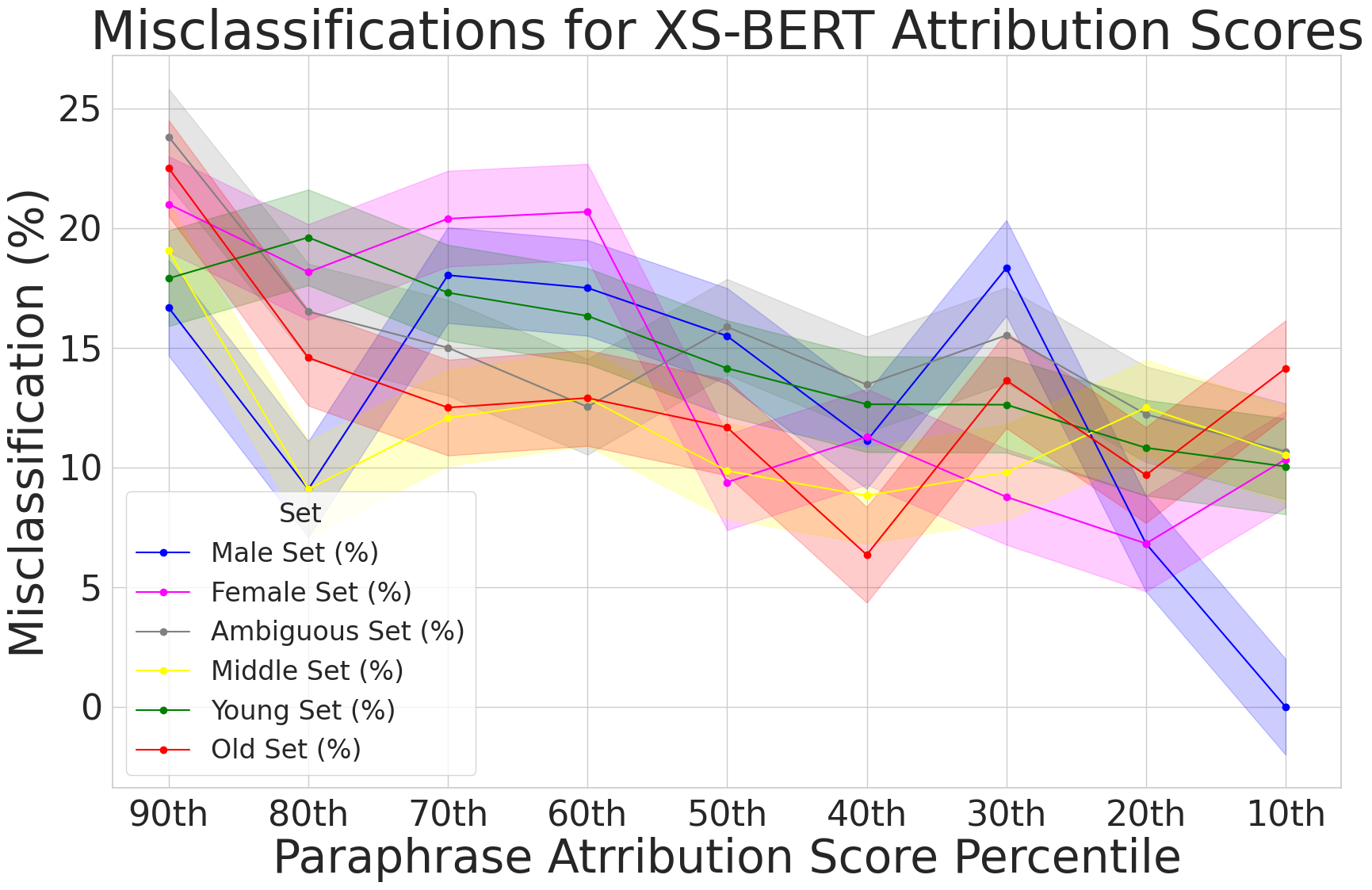}
\caption{XSBERT attribution percentile misclassifications. Lower model attributions (90\textsuperscript{th} percentile score) exhibit higher misclassification rates, indicating that reduced model confidence aligns with poorer paraphrase accuracy.}
\label{fig:xsbert_mis_perc}
\end{figure}

We examine the performance of LLMs on paraphrased sets in comparison to the semantic attributions between original and paraphrased sentences. We rely on explainability methods that enhance our understanding of model decisions, which are obtained with the integration of XSBERT \cite{moeller-etal-2023-attribution} and DistilRoBERTa \cite{sanh2019distilbert}. 
XSBERT is an adaptation of the Siamese network architecture applied to sentence transformers. It simultaneously processes paired inputs and maps them onto a scalar output, which is represented as model attribution scores and attribution errors for an input sentence pair. 
Traditional methods like cosine similarity are effective for determining the likeness of the sentences overall but do not account for token-wise attributions. In contrast, XSBERT utilizes integrated Jacobians which provide a richer and matrix-based attribution that measures the strength of relationships and the specific features contributing to model decisions. The trends seen in Figure \ref{fig:xsbert_mis_perc} suggest that the explainability of paraphrasing plays a crucial role in the model's assessment of the context.

\section{Two-shot Baselines}
In Table \ref{table:baselines}, we observe some of the existing 2-shot baselines taken from BIG-Bench with popular LLMs. The reported values are an average of three runs of the SIQA test set and paraphrased sets. Compared to the existing baselines, \texttt{LLAMA2} performs significantly better. It is also worth noting that with our proposed answer selection approach using log-likelihoods, we see an improvement over the officially reported baseline scores with \texttt{LLAMA2}. 

\begin{table}[t]
    \centering
    \small
    \begin{tabular}{|l|c|c|}
        \hline
        \textbf{Model} & \textbf{Accuracy} & \textbf{Config} \\
        \hline
        Random & 33.3 & - \\
        GPT-3 Small & 33.7 & 2 shot \\
        GPT-3 XL & 35.5 & 2 shot \\
        GPT-3 200B & 46.8 & 2 shot \\
        PaLM 64B & 58.0 & 2 shot \\
        \texttt{LLAMA2} (original) & 50.4 & 0 shot \\
        \texttt{LLAMA2} (ours) & 57.8 & 0 shot \\
        \texttt{MISTRAL} & 66.3 & 0 shot \\
        \hline
    \end{tabular}
    \caption{SIQA 2-shot Baselines from BIG-Bench.}
    \label{table:baselines}
\end{table}

\section{Optimal LLaMA Hyperparameters}

We describe our experimental setup configurations for optimal usage of \texttt{LLAMA2} and \texttt{MISTRAL} for paraphrase generation and QA inference.\footnote{The weights in the GPTQ format which is supported by Llama.cpp are available in the huggingface model registry at TheBloke/Llama-2-13B-chat-GPTQ}

\begin{table}[t]
\centering
\small
\begin{tabularx}{\columnwidth}{|l|X|}
\hline
\textit{Model} & Paraphrase generation - LLAMA2-chat (13B) \\
&  QA Evaluation - LLAMA2-chat (13B), MISTRAL-instruct (7B) \\ \hline
\textit{Dataset} & SIQA Validation Set (1954 samples) \\ \hline
\textit{Implementation} & llama.cpp (Python bindings) \\ \hline
\textit{GPU} & NVIDIA Tesla P100 SXM2 \\ \hline
\textit{Memory Size} & 16GB \\ \hline
\textit{Weights} & GPTQ with 6-bit quantization \\ \hline
\textit{Temperature} & 0.8 \\ \hline
\textit{Repetition Penalty} & 1.1 \\ \hline
\textit{Context Length} & 4096 \\ \hline
\end{tabularx}
\caption{Experiment setup for LLM generation and inference.}
\label{tab:hyperparameters}
\end{table}

\begin{table}[t]
    \centering
    \small
    \begin{tabular}{|c|c|c|}
        \hline
        \textbf{Temperature} & \textbf{RP} & \textbf{LexHub alignment} \%(\(M < F\))\\
        \hline
        0.4 & 1.1 & 64\% \\
        \hline
        0.6 & 1.1 & 65\% \\
        \hline
        0.8 & 1.1 & 72\% (default) \\
        \hline
        1.0 & 1.1 & 70\% \\
        \hline
        0.8 & 1.18 & 68\% \\
        \hline
        0.8 & 1.0 & 61\% \\
        \hline
        0.8 & 0.9 & 59\% \\
        \hline
        0.8 & 0.8 & 53\% \\
        \hline
    \end{tabular}
    \caption{Various values of temperature and repetition penalty for paraphrase generation. We select the optimal hyperparameters considering the best LexHub alignment.}
    \label{table:hyperparameter_generation}
\end{table}

\section{Evaluation Formula}
In the context of multiple-choice question answering, a common approach used with LLMs to select the MCQ choice is standard text completion followed by exact string matching. However, this approach encounters difficulties in scenarios with multiple matches or no matches for the target answers. We instead use a more structured approach where logits for appended target answers are calculated. We compute the likelihood of each answer option and then select the one with the highest likelihood as the answer. We use the \texttt{LLAMA2} model in `echo mode` to repeat the input sequence appended with the completion tokens. Additionally, we set the 'max\_tokens=1' to ensure that the model halts after one generated token after evaluating the prompt, and a 'presence\_penalty' of 0 is employed to avoid penalizing the model for repeating a token present in the prompt.

The mathematical formulation for selecting the answer is as follows:

\[
\texttt{answer} = \arg \max_{i \in \{1, 2, 3\}} \log P(\texttt{choice}_i | \texttt{context})
\]

Here, $\log P(\texttt{choice}_{i} | \texttt{context})$ represents the log-likelihood of the probability of each choice \(i\) given the context.

\section{Demographic Artefacts in Paraphrases}
 We analyze the text of the paraphrases for recurring themes, slang, modern colloquialisms, or specific stylistic choices that distinguish these paraphrases from those targeted at other demographics. For a detailed view of the distinct words and their frequencies, refer to Table \ref{table:distinct_words}. The analysis of the most common tokens in the Young demographic paraphrases reveals a mix of generic English words along with some interesting patterns that may hint at expressions and artifacts typical of younger audiences. Here are some noteworthy findings:

\begin{table*}[t]
    \centering
    \small
    \setlength{\tabcolsep}{5pt}
    \begin{tabular}{|c|c|c|c|c|c|}
        \hline
        \multicolumn{3}{|c|}{\textbf{Demographic Set - Age}} & \multicolumn{3}{|c|}{\textbf{Demographic Set - Gender}} \\
        \hline
        \textbf{Middle-aged} & \textbf{Old} & \textbf{Young} & \textbf{Ambiguous} & \textbf{Female} & \textbf{Male} \\
        \hline
        close (31) & young (59) & totally (753) & individual (52) & gently (43) & buddies (105) \\
        \hline
        thoughtful (27) & acquaintances (54) & super (156) & leaving (48) & herself (39) & buddy (80) \\
        \hline
        own (26) & lovely (53) & sick (149) & experienced (41) & adorable (35) & hand (28) \\
        \hline
        late (25) & dear (49) & total (75) & resulting (39) & beautiful (29) & stepped (28) \\
        \hline
        looking (24) & delightful (42) & know (64) & social (34) & money (28) & headed (28) \\
        \hline
    \end{tabular}
    \caption{Distinct words and their frequencies in each category.}
    \label{table:distinct_words}
\end{table*}

\begin{itemize}
\item Words such as "totally" rank very high in frequency, indicating a preference for expressive and emphatic language characteristic of informal, youthful speech
\item The frequent use of pronouns ("you", "her", "his", "they") and conversational verbs ("want", "feel", "like") suggests a conversational and relational focus in the paraphrasing style
\item Slang and Colloquialisms, eg,
\textit{"Alex totally bailed on their stressful life and headed to Mexico for some serious chillaxation and eye-candy sightseeing."}
The use of "totally bailed" and "chillaxation" (a blend of "chill" and "relaxation") alongside "eye-candy" reflects a casual, slang-rich language style.

\end{itemize}

The older demographic favors a style that leans towards formality, with clear and precise language. This includes the use of proper grammar and punctuation, as well as a preference for more traditional vocabulary over trendy terms or slang. Furthermore, their style reflects a preference for storytelling narrative expressions or a more comprehensive approach to communication. For example, in the phrase \textit{"Sydney ventured out for a \textbf{spooky Halloween excursion}, accompanied by \textbf{his delighted companions}."}, the highlighted part employs a narrative style that describes the action and its effect.

Finally, in the Gender Ambiguous demographic group, the significant presence of "their" underscores an intentional use of gender-neutral language. This shows a focus on inclusivity and avoiding gender-specific references.

\section{Semantic Clusters for Paraphrases}

\begin{figure}[t]
\begin{center}
    \includegraphics[scale=0.40]{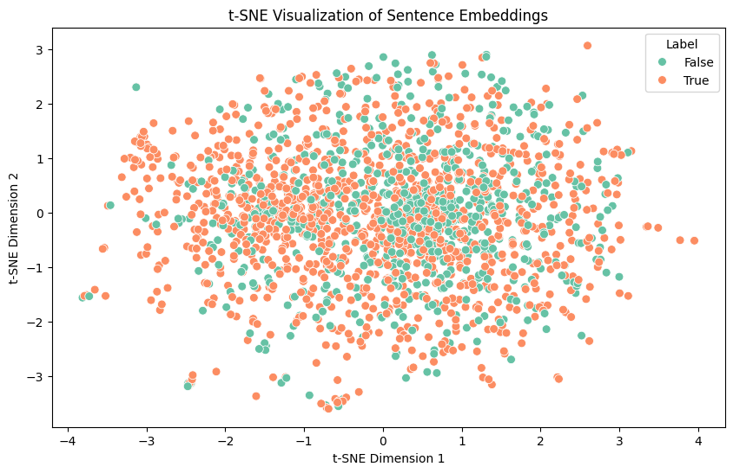}
\end{center}
\caption{t-sne clustering of aligned vs unaligned paraphrase embeddings}
\label{fig:semantic_cluster}
\end{figure}

In Figure \ref{fig:semantic_cluster} we explore the t-SNE clustering of aligned versus unaligned paraphrase embeddings. With this, we aim to visualize whether there is a distinct separation in the embeddings of paraphrase sets that disagree compared to those that agree. The plots reveal no significant divergence between the embeddings of paraphrases in agreement and those in disagreement. This suggests that while semantic discrepancies may contribute to LexHub's agreement, the differentiation is not markedly distinct.  The outcome implies that factors beyond mere semantic content may have a greater influence on the alignment process

\begin{figure}[t]
\begin{center}
    \includegraphics[scale=0.155]{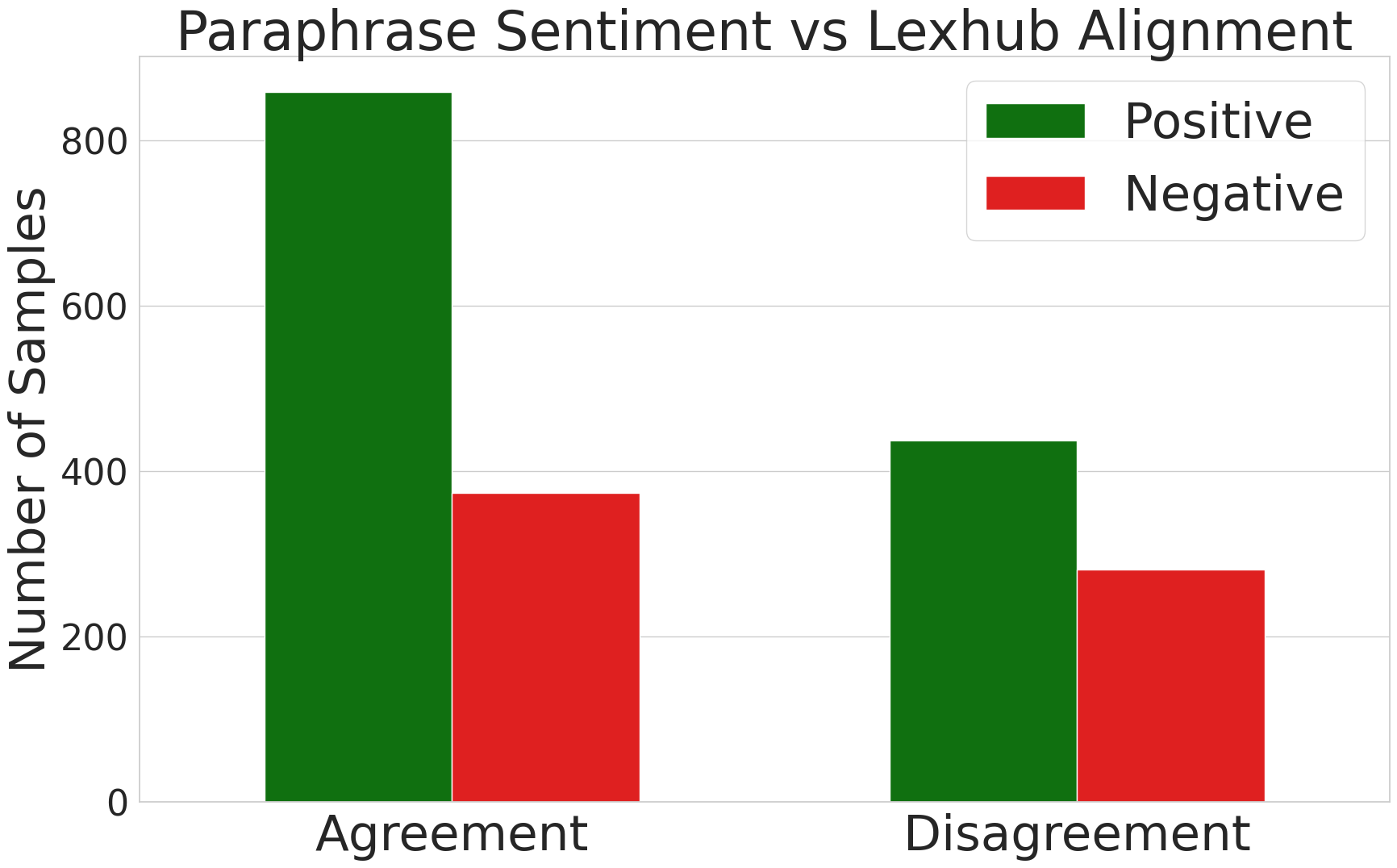}
\end{center}
\caption{VADER sentiment score of male and female paraphrases compared with their alignment with LexHub. A positive label indicates the sentiment score of the female paraphrase being greater than the male counterpart and vice versa}
\label{fig:lex_alignment}
\end{figure}

\begin{figure}[t]
\includegraphics[scale=0.165]{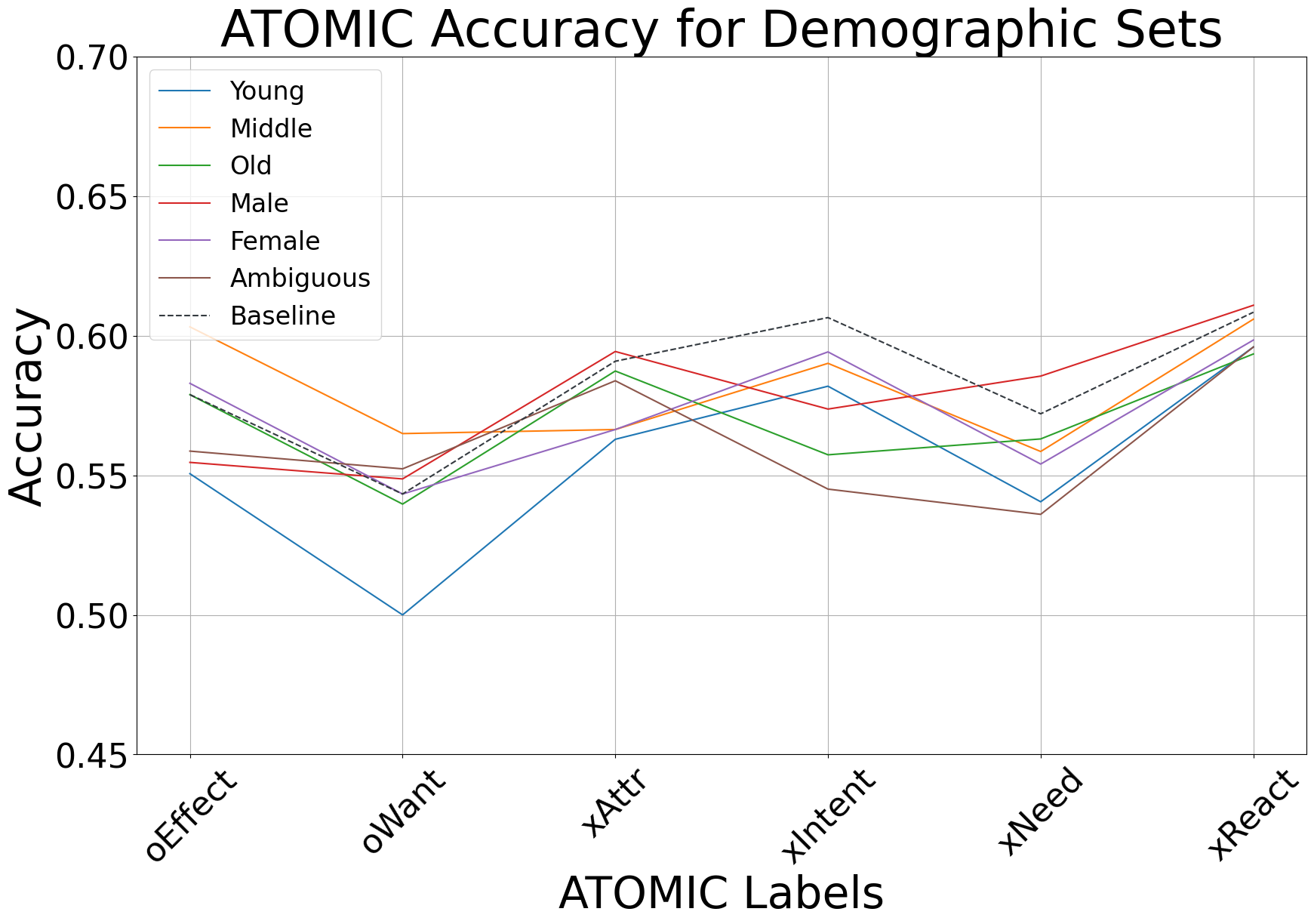}
\caption{SIQA Atomic Accuracy}
\label{fig:atomic_accuracies}
\end{figure}

\noindent
\section{Sentiment Alignment with LexHub} 
Our analysis incorporates sentiment scores of the paraphrases derived from a lexicon and rule-based sentiment analysis tool, NLTK VADER \cite{hutto2014vader}. 
With this, we aim to discern whether the sentiments of Male paraphrases differ significantly from those of Female paraphrases and how these differences align with LexHub's categorizations of "agreement" and "disagreement". 

In Figure \ref{fig:lex_alignment}, our findings reveal a notable pattern: in paraphrase sets categorized by LexHub as being in "agreement", 33.8\% exhibit a lower sentiment score for Female style paraphrases compared to the Male style. Conversely, in sets classified under "disagreement", this is 43.0\%.
This indicates a more pronounced sentiment disparity in paraphrases that fail to align according to LexHub standards, suggesting that also the sentiment plays a role in the alignment process. When the paraphrase sentiment follows the observed Female or Male linguistic style, the alignment with LexHub is higher.

\section{Performance on ATOMIC questions}
ATOMIC knowledge is categorized into social interactions, physical events, and mental states, with dimensions such as causes (xIntent, xNeed), effects (xEffect, oEffect), and attributes (xAttr, oReact). To label the SIQA development set with ATOMIC labels, we manually map the social situations and behaviors described in SIQA to the corresponding dimensions and categories found in the ATOMIC dataset.
The analysis of ATOMIC label interpretation across demographics within the SIQA dataset reveals distinct patterns in how LLM predictions are affected when different groups process social scenarios, shown in Figure \ref{fig:atomic_accuracies}. For example, younger individuals often struggle with xNeed, which is reflective of their ability to describe their emotional states. Middle-aged paraphrasing improves the interpretation of social cues in xIntent and xReact. The impact of paraphrasing is not uniform across all ATOMIC categories. For instance, xNeed (needs before an action) and oWant (desires post-event) often see greater fluctuations in accuracy. This could be because the subtleties of needs and desires are sensitive to linguistic differences. The paraphrased datasets generally show close accuracy in recognizing xAttr (attributes of individuals) and xReact (reactions to social situations), implying that describing attributes and predicting reactions is not extremely dependent on language style.

\label{sec:appendixF}

\end{document}